\documentclass{ecai} 

\usepackage{latexsym}
\usepackage{amssymb}
\usepackage{amsmath}
\usepackage{amsthm}
\usepackage{booktabs}
\usepackage{enumitem}
\usepackage{graphicx}
\usepackage{color}

\usepackage{url}

\usepackage{tikz}
\usepackage{pgfplots}
\pgfplotsset{compat=1.18}
\usepackage{comment}

\usepackage{silence}
\usepackage{styles/newanalogy}
\usepackage{styles/cjkex}

\newtheorem{theorem}{Theorem}

\newtheorem{corollary}[theorem]{Corollary}

\newtheorem{definition}{Definition}
\newtheorem{remark}{Remark} 

\makeatletter
\newtheorem*{rep@theorem}{\rep@title}
\newcommand{\newreptheorem}[2]{%
\newenvironment{rep#1}[1]{%
    \def\rep@title{#2 \ref{##1}}%
    \begin{rep@theorem}}%
    {\end{rep@theorem}}}
\makeatother

\newreptheorem{theorem}{Theorem}

\newcommand{\BibTeX}{B\kern-.05em{\sc i\kern-.025em b}\kern-.08em\TeX}

\newcommand{\Retoile}{\mathbb{R}^*}
\newcommand{\Retoileplus}{\mathbb{R^{+}}\setminus\{0\}}
\newcommand{\moyp}[3]{\sqrt[#1]{\displaystyle\frac{1}{2}({#2}^{#1}+{#3}^{#1})}}
\newcommand{\moygen}[4]{\lim_{#1\rightarrow #2} \moyp{#1}{#3}{#4}}
\newcommand{\graybox}[1]{\fcolorbox{gray!30}{gray!30}{#1}}

\newcommand{\ie}{\emph{i.e.}}
\newcommand{\eg}{\emph{e.g.}}

\begin{document}

\begin{frontmatter}


\paperid{58} 


\title{
Any four real numbers are on all fours with analogy
}


\author[A]{\fnms{Yves}~\snm{Lepage}\orcid{0000-0002-3059-4271}\thanks{Corresponding Author. Email: yves.lepage@waseda.jp.}}

\author[B,C]{\fnms{Miguel}~\snm{Couceiro}\orcid{0000-0003-2316-7623}}
\address[A]{Waseda University, Japan}
\address[B]{University of Lorraine, CNRS, LORIA, France}
\address[C]{INESC-ID, IST, Universidade de Lisboa, Portugal}

\begin{abstract}
This work presents a formalization of analogy on numbers that relies on generalized means.
It is motivated by recent advances in artificial intelligence and applications of machine learning, where the notion of analogy is used to infer results, create data and even as an assessment tool of object representations, or embeddings, that are basically collections of numbers (vectors, matrices, tensors).
This extended analogy use asks for mathematical foundations and clear understanding of the notion of analogy between numbers. 
We propose a unifying view of analogies that relies on  generalized means defined in terms of a power parameter. 
In particular, we show that any four increasing positive real numbers is an analogy in a unique suitable power.
In addition, we show that any such analogy can be reduced to an equivalent arithmetic analogy and that any analogy equation has a solution for increasing numbers, which generalizes without restriction to complex numbers.
These foundational results provide a better understanding of analogies in areas where representations are numerical. 
\end{abstract}
\end{frontmatter}

%

\section{Introduction}
\label{section:introduction}

\paragraph{Background and earlier work.}
In artificial intelligence, the terms analogical inference or analogical reasoning are often used.
Related work focuses on the study of relationships between two pairs of objects $A$ and $B$ on the one hand, and of objects $C$ and $D$ on the other. 
There are several possible uses for such a configuration.
Typically, one may want to judge whether the relationship between $C$ and $D$ is the same as that between $A$ and $B$.
The quality of the similarity of such relationships can be then assessed, and one can discuss attribute or relationship similarity~\citep{turney_compling_06},
following the foundational work by Gentner~\citep{gentner_structure_mapping_1983}.
One may also see $A$ and $C$ as problems,
$B$ as a solution to problem $A$,
and ask whether the transposition
of the ratio of $A$ to $B$ on $C$ generates a $D$, and
to what extent the generated $D$ is a solution to problem $C$.
This setting is at the core of case-based reasoning~\citep{BadraL22,OursFadi,LieberNP21}. 

The underlying principle is that of \emph{analogical inference}, and it has been integrated into various machine learning tasks such as preference learning and recommendation~\citep{FahandarH18,FahandarH21,Mitchell21} and, more generally, in classification \citep{classification}.
In addition, analogical extrapolation (inference) can solve difficult reasoning tasks such as tests of academic ability or visual question answering~\citep{DafnaSaaai,PeyreSLS19,SadeghiZF15} or checking the meaning of a target sentence~\citep{TSV}.
It can also support dataset augmentation  through analogical extension and extrapolation~\citep{couceiro3} and guide computational creativity \cite{Goel19}.
Also, analogical transfer can be realized by transfer learning~\citep{CornuejolsMO20,alsaidiTrans} where the idea is to take advantage of what has been learned in a source domain  to improve the learning process in a target domain related or linked to the source domain.
Finally, analogy creation can provide useful explanations that build on facts or counterfactuals~\citep{Hullermeier20} and  guide counterfactual generation~\citep{KeaneS20}.

In particular domains such as machine translation,
this type of analogical reasoning has been used~\citep{nag_ex_eng,lan_eacl_09} and
the connection with case-based reasoning is well recognized~\citep{collins_ebmt_2003}.
Still in natural language processing, analogy-based approaches have been used for the tasks of morphological analysis~\citep{rhouma-langlais:2014:Coling,rhouma_iccbr_2018} or word generation~\citep{fam_jetai_2022}.
For instance,
\citep{murena_ijcai_2020} proposes to minimize the algorithmic complexity of the program describing the ratio of $A$ to $B$,
\ie, the transformation of $A$ into $B$,
and then apply the program of minimum complexity to $C$ to produce $D$, a new word, yielding high performance on a set of several million morphological analogies in a dozen languages.

But analogical reasoning is not exactly analogy.
In everyday life, 
the English word {\it analogy} easily takes on the meaning of ``reasoning by analogy'' or even ``simple comparison'', and is quite often linked to fallacious reasoning.
The present article is about mathematical analogy, 
which is not quite that.
It is a relationship on a quadruple that does not privilege any particular ratio,
for instance, that of~$A$ to~$B$ in preference to that of~$C$ to~$D$,
nor does it favor a particular direction,
for instance, from~$A$ to~$B$ in preference to~$B$ to~$A$,
nor does this view favor one term over the others,
even if they are not all interchangeable.
The properties of interchangeability, which will be recalled in Sections~\ref{subsection:conformity} and~\ref{subsection:eight:equivalent:forms},
have already been the subject of observations and debate since antiquity, see
\eg, the Encyclopédie~\citep{dumarsais_analogie_1771}.

\paragraph{Motivation and recent work.} The recent increasing interest in analogies and analogical reasoning was partially due to the successes  of deep learning together with distributional representations (embeddings). For instance, \cite{DrozdGM16,MikolovYZ13} show  that vector representations of quadruples respecting certain linear transformations satisfy common properties of analogies, whereas  \cite{SchnabelLMJ15} unveils the potential analogies as a benchmark to evaluate the quality of embedding models. This exploration has extended to complex structures such as knowledge graphs (KG) over multimodal domains, to address tasks such as named entity recognition, link prediction, relation discovery (abduction) and KG bootstrapping and completion \cite{UshioASC20,LDC23,JarnacCM23}, by leveraging multimodal knowledge embeddings \cite{TrouillonWRGB16,XieLLS17,WangLLZ19,WangWYZCQ21}.   Despite these impressive results by such analogy-based approaches in rather complex tasks, many works have questioned the retrieved analogical relations, their dependency and limitation with respect to underlying representation model, and even the evaluation procedure \cite{Linzen16,Schluter18}. Many works have been advocating for foundational mathematical frameworks and experiments to gain a better understanding of the analogical capabilities of embedding models as well as recent large language models \cite{GladkovaDM16,BouraouiJS18,AllenH19,PetersenP23}. 

This contribution goes in the latter direction. More precisely, we go back to the very foundations of analogies, and propose a well-defined and unifying framework for numerical analogy. In doing so, we enable its application to individual dimensions of vector representations, and hence open avenues for a better understanding and justified use of analogy in machine learning downstream tasks.

Our proposal is rooted in the intuitive and classical idea of an analogy: we say that $(A,B,C,D)$ constitutes a valid analogy, denoted by $\analogy{A}{B}{C}{D}$, if there is a $p\in \mathbb{R}$, such that the $p$-generalized mean of the \emph{extremes} $A$ and $D$ is equal to the $p$-generalized mean of the \emph{means} $B$ and $C$ (see Section~\ref{section:means:generalized}). 
One of the advantages of relying on this parameterized notion on $p$, is that it naturally subsumes well known mean notions, such as the commonly used  arithmetic (for $p=1$) and geometric (for $p=2$) means.
Endowed with this notion, we introduce the notion of {\it analogy in power~$p$} and study its properties such as reflexivity, symmetry and transitivity of conformity (the symbol $::$, see below), as well as properties of interchangeability of terms.

\paragraph{Main results.} The main contributions of this article can be summarized in three rather surprising results.
The first one states that analogies exist and are unique between any quadruple of positive real numbers
(the numbering of the results does not reflect the importance but the order afterwards).
\begin{reptheorem}{main:result:analogy:reals}
Given four positive real numbers,
all different,
we can always see an analogy between them
and this analogy is unique.
\end{reptheorem}

In fact, this result extends to any quadruple of co-linear complex numbers on the same side of $0$ on the line. 

The most common analogy between numbers, arithmetic analogy, is just a particular case for $p = 1$. In addition to Theorem~\ref{main:result:analogy:reals}, our second main result states that any analogy between four positive reals can be thought of as an arithmetic analogy. 

\begin{reptheorem}{main:result:reduction:analogy:arithmetic}
Any analogy in $p$ between four positive real numbers can be
reduced to an arithmetic analogy.
\end{reptheorem}

As vector representations in machine learning make use of real numbers, the main results are presented with respect to real numbers, but extensions are possible. In particular, our third main result states that every analogical equation of the form $\analogy{A}{B}{C}{x}$ is solvable over the complex numbers.

\begin{reptheorem}{main:result:solving:complex}
Conditioned on $p$, it is possible to solve any analogical equation with complex terms.
\end{reptheorem}


\noindent We now set the notation and terminology used throughout the article.

\paragraph{Notation and terminology.} We use capital letters $A$, $B$, $C$ and $D$ when referring to {\it terms} of an analogy in general, and use the letters $a$, $b$, $c$ and $d$ when these terms are numbers. 
We classically denote with $\mathbb{R}$, $\Retoile$, $\Retoileplus$ and $\mathbb{C}$ the sets of real numbers, real numbers without $0$, positive numbers and complex numbers.

An analogy is classically denoted by $\analogy{A}{B}{C}{D.}$ 
The symbol~$:$ is that of \emph{ratio}.
We read the symbol~$::$ \emph{conformity} to avoid the terms equality or identity, which are too closely linked to a notion of equivalence relation.
We will read \emph{conformity in~$p$} the symbol~$::^p$ introduced later below.

\begin{table*}[ht]
\begin{center}
\begin{tabular}{clc@{$\;\;\Leftrightarrow\;\;$}c@{$\;\;\Leftrightarrow\;\;$}c@{=}c}
	& $b$ as means  \\
	$\analogy[][=]{a-b}{b-c}{a}{a}$ & arithmetic & $a-b : b-c = 1$ & $b = \displaystyle\frac{1}{2}(a + c)$ & $b$ & $\displaystyle\frac{1}{2}(a + c)$ \\[2ex]
	$\analogy[][=]{a-b}{b-c}{a}{b}$ & geometrical & $\analogy[][=]{ab-b^2}{ab-ac}{ab}{ab}$ & $b^2 = ac$ & $\log{b}$ & $\displaystyle\frac{1}{2}(\log{a} + \log{c})$ \\[2ex]
	$\analogy[][=]{a-b}{b-c}{a}{c}$ & harmonic  & $\analogy[][=]{ca-cb}{ab-ac}{ac}{ac}$ & $b = \displaystyle\frac{2ac}{a + c}$ & $\displaystyle\frac{1}{b}$ & $\displaystyle\frac{1}{2}(\frac{1}{a} + \frac{1}{c})$ \\
\end{tabular}
\end{center}
\caption{%
    \label{tab:medietes}
    Some ``medieties'' from~\citep{michel_persee_1949}.
    With division as the ratio,
    we obtain the three classic means,
        arithmetic,
        geometric and
        harmonic
    from analogies whose first three terms are the same.
    Only the fourth term varies and successively takes the values $a$, $b$ and $c$.
}
\end{table*}

\section{Generalized means and analogy}
\label{section:means:generalized}

For a quick historical reminder on the mathematical notion of analogy in Greek antiquity, and to show the link to the notion of mean, let us recall that, according to some scholars,
the Greek word for analogy
(``again the (same) ratio'') might have emerged after a period of indecision over how to refer to continuous analogy,
\ie, $\analogy{A}{B}{B}{D}$
(note the repetition of $B$)
and to discrete analogy,
\ie, $\analogy{A}{B}{C}{D.}$
In the Nicomachean Ethics~\citep[1131 a29 -- 1131 b2]{aris_eth} Aristotle repeats that there are two types of analogy
each called discrete and continuous respectively,
the first involving four terms,
the second involving three terms, one of which is repeated.
\begin{quote}
``Analogy is an equality of ratios and comes in four terms at least.
It is obvious that the discrete one is indeed in four terms.
But so is the continuous one; [...] B is uttered twice [...]''
\end{quote}

In the case of a continuous analogy,
with division as a ratio,
$b$ can be calculated in function of $a$ and $d$:
\[
\analogy[\,\textbf{$\div$}\,][=]{a}{b}{b}{d}
\;\; \Leftrightarrow \;\;
b^2 = a \times d
\;\; \Leftrightarrow \;\;
b = \sqrt{a \times d}.
\]
This formula for $b$ is that of the geometric mean of $a$ and $d$.

If the ratio is subtraction,
then, again for a continuous analogy :
\[
\analogy[\,-\,][=]{a}{b}{b}{d}
\;\; \Leftrightarrow \;\;
2 \times b = a + d
\;\; \Leftrightarrow \;\;
b = \frac{1}{2}(a + d)
\]
Here, $b$ is the arithmetic mean of $a$ and $d$.

From the above, it is clear that continuous analogy is intimately linked to the notion of a mean.

The Pythagoreans in the Antiquity and mathematicians of the Middle Ages have studied the links between the various conceivable means and analogy.
Particular mention should be made of the studies on ``medieties''
(lat. medietas or mediocritas, cf. \citep{michel_persee_1949}).
Table~\ref{tab:medietes} illustrates one way, among others, of extracting the three arithmetic, geometric and harmonic means from similar, but discrete, analogies.

%

The generalization of the notion of a mean,
unrelated to analogy,
beyond arithmetic, geometric or harmonic means,
was given in 1882 in an article by Hölder~\citep{hoelder_grenzwerthe_1882}.
The aim of the present article is to link the mathematical notion of analogy to this generalization of the notion of mean.

\subsection{Definition of generalized means}

The generalized mean of several real positive numbers $x_1,\ldots x_N$ is the value
\[
m_p(x_1,\ldots x_N) =
\lim_{r\rightarrow p} \sqrt[r]{\frac{1}{N} \sum_{i=1}^{N}x_i^r}
\]
for all $p \in \; ( -\infty, +\infty)$.
In particular, we find:
\begin{itemize}
    \item the arithmetic mean for $p =1$;
    \item the harmonic mean for $p =-1$;
    \item the root mean square for $p =2$;
    \item the geometric mean when $p$ tends to $0$:
$$lim_{p\rightarrow 0}m_p(x_1,\ldots x_N) = \sqrt[N]{\prod_{i=1}^{N} x_i}.$$
\end{itemize}
Finally, when $p$ tends towards $+\infty$, it is the maximum of the numbers, $\max(x_1,\ldots x_N)$,
and
when $p$ tends towards $-\infty$, it is the minimum of the numbers, $\min(x_1,\ldots x_N)$.

\begin{remark}
This generalizes to the case when $a$, $b$, $c$, $d$ and $p$ are complex numbers, except, \eg, in the undefined cases  $0^p$ for $p<0$.
\end{remark}

\subsection{Specialization to two real numbers}

The generalized mean of two numbers $a$ and $d$ is the value
\[
m_p(a, d) =
\lim_{r\rightarrow p} \sqrt[r]{\frac{1}{2} (a^r + d^r)}
\]
for all $p \in \mathbb{R}$,
and even for $p$ equal to $-\infty$ or $+\infty$.
Figure~\ref{figure:means:generalized} illustrates the curve obtained for the particular values $a=2$ and $d=5$.

It is tempting to think that the generalized mean function of two numbers would be a function
that behaves like an odd function with respect to a particular value of $p$.
But this is not usually the case. 
Figure~\ref{figure:means:generalized} illustrates this.

Before turning to the results, we would like to point out that, in the following, we shall not go into detail on all the demonstrations, as they always adopt the same structure.
In the general case where $p$ is neither 0 nor infinite,
the limit is not necessary,
neither is taking the $p$-th root,
the one-half factor can be eliminated
and demonstrations can then directly exploit the formula $a^p + d^p$ or the equality $a^p + d^p = b^p + c^p$.
In the case $p = 0$, the formulas of the geometric mean are used: $a \times d$ or $a \times d = b \times c$.
Finally, in the two infinite cases, formulas with $\min$ and $\max$ are used.
We will also omit some proofs, but they are to be found in the Supplementary material.


\subsection{Definition of analogy through generalized means}
\label{section:analogy:mean}
\label{subsection:generalized:analogy:definition}

We can now define analogies in terms of generalized means.

\begin{definition}
\label{definition:generalized:analogy}
On four positive real numbers, 
we define \emph{analogy in power~$p$}
as follows:
\begin{align*}
	\analogy[][::^p]{a}{b}{c}{d}
		& \; \overset{\text{def.}}{\Leftrightarrow} \;
			m_p(a, d) = m_p(b, c) \\
		& \; \Leftrightarrow \;
			\moygen{r}{p}{a}{d}
				=
			\moygen{r}{p}{b}{c} ~.
\end{align*}
\end{definition}

Figure~\ref{figure:means:generalized} 
indicates the values of the generalized means for $a = 2$ and $d = 5$, according different values of $p$ on the x-axis.
We see the harmonic ($2.86$), the geometric ($\sqrt{2 \times 5} \simeq 3.16$), the arithmetic ($3.5$) and the quadratic ($3.81$) mean values. Similarly, the limits at $-\infty$ and $+\infty$ are the minimum and maximum, $2$ and $5$ respectively.  Note that the curve is not symmetrical w.r.t. any horizontal line.

\begin{figure}
\begin{tikzpicture}
\begin{axis}
	[
		xmin = -100, xmax = 100,
		ymin = 1, ymax = 6,
		xlabel = {$p$},
		ylabel = {$y = \text{mean}_p(2,5)$},
	]
	\addplot[thick, domain=-100:-0.01] {(0.5 * (2 ^ x + 5 ^ x)) ^ (1/x)};
	\addplot[thick, domain=0.01:100] {(0.5 * (2 ^ x + 5 ^ x)) ^ (1/x)} ;
	
	\addplot[dotted, domain=-100:100] {2} node [pos=0.12, anchor=south, below] {\small minimum} ; 
	\addplot[dotted, domain=-100:-1] {2.86} node[right] {\small harmonic} ; 
	\addplot +[mark=none, black, dotted] coordinates {(-1, 0) (-1, 2.86)};
	\addplot[dotted, domain=-100:0] {3.16} node[right] {\small geometric} ; 
	\addplot +[mark=none, black, dotted] coordinates {(0, 0) (0, 3.16)};
	\addplot[dotted, domain=-100:1] {3.50} node[right] {\small arithmetic} ; 
	\addplot +[mark=none, black, dotted] coordinates {(1, 0) (1, 3.50)};
	\addplot[dotted, domain=-100:2] {3.81} node[right] {\small quadratic} ; 
	\addplot +[mark=none, black, dotted] coordinates {(2, 0) (2, 3.81)};
	\addplot[dotted, domain=-100:100] {5} node[pos=0.88, anchor=north, above] {\small maximum} ; 
	\end{axis}
\end{tikzpicture}
\caption{\label{figure:means:generalized}%
 Visualization of generalized means of $a=2$ and $d=5$.    
}
\end{figure}

In words,
four positive real numbers $a$, $b$, $c$ and $d$ are in analogy,
if and only if
there exists a $p$
such that
the generalized mean in $p$ of the extremes $a$ and $d$
is equal to
the generalized mean in $p$ of the means $b$ and $c$.
Observe that the definition puts the extremes and the means on each side of the equality sign.
This slightly differs from usual understanding of analogy, clearly visible from the notation, that ratio is instrumental in analogy. For instance the standard intuitive interpretation of arithmetic analogy is $a - b = c - d$ whereas the above definition states that $a + d = b + c$, which is fortunately equivalent. 
This definition raises several questions:
firstly, under what conditions
we can find a $p$ such that the analogy holds
and,
secondly, to determine this $p$.

Before answering these questions, however, we need to check that the definition given above corresponds to the usual perception of a mathematical analogy.
For that, we need to examine whether we have reflexivity and symmetry for $::$ and whether the eight equivalent forms of an analogy are verified here.\footnote{Transitivity is not usually required for $::$ in general; a simple relation of resemblance, not equivalence, suffices.
Hence our insistence on calling $::$ conformity and not equality or identity.
}

\subsection{Reflexivity, symmetry and transitivity of $::^p$}
\label{subsection:conformity}

For the reflexivity of $::^p$,
it is trivial to note that,
for any two positive real numbers $a$ and $b$,
for any $p$,
we always have
\begin{equation}
\label{eq:reflexivity:conformity}
    \analogy[][::^p]{a}{b}{a}{b},
\end{equation}
\ie,
\begin{equation}
\label{eq:reflexivity:conformity:explanation}
	\moygen{r}{p}{a}{b}
		=
	\moygen{r}{p}{b}{a}.
\end{equation}

The symmetry of $::^p$ is expressed as follows:
\[
	\analogy[][::^p]{a}{b}{c}{d}
		\Leftrightarrow
	\analogy[][::^p]{c}{d}{a}{b}.
\]
It is verified because:
\begin{align*}
	\analogy[][::^p]{a}{b}{c}{d}
	& \Leftrightarrow
		m_p(a, d)
			=
		m_p(b, c)
    \\
	& \Leftrightarrow
		m_p(b, c)
			=
		m_p(a, d)
    \\
	& \Leftrightarrow
		m_p(c, b)
			=
		m_p(d, a)
    \\
	& \Leftrightarrow
		\analogy[][::^p]{c}{d}{a}{b}.
\end{align*}

Let us repeat that transitivity is not generally necessary for analogy.
It just happens to exist for $::^p$, $p$ different from~0 and infinity.
This can be stated as:
\begin{equation}
	\left\{
		\begin{aligned}
			\analogy{a}{b}{c}{d} \\
			\analogy{c}{d}{e}{f}
		\end{aligned}
	\right.
		\quad\Rightarrow\quad
	\analogy{a}{b}{e}{f}.
\end{equation}

For the general case where $p$ is in $(-\infty; +\infty)$,
but different from~0,
transitivity holds.
We give the proofs and the details for the cases for infinity in the Supplementary Material.

\subsection{Eight equivalent forms of analogy}
\label{subsection:eight:equivalent:forms}

Classical formalizations of mathematical analogy allow eight equivalent forms to write the same analogy, by playing on the position of the terms in the analogy.
\begin{align*}
\analogy{a}{b}{c}{d} \qquad\ \analogy{c}{a}{d}{b} \\
\analogy{a}{c}{b}{d} \qquad\ \analogy{c}{d}{a}{b} \\
\analogy{b}{a}{d}{c} \qquad\ \analogy{d}{b}{c}{a} \\
\analogy{b}{d}{a}{c} \qquad\ \analogy{d}{c}{b}{a}
\end{align*}
The article Proportion in the Encyclopédie~\citep{rallier_proportion_1771} mentions the two best-known:
\emph{invertendo}, the inversion of ratios $\analogy{b}{a}{d}{c}$
and
\emph{permutando}, the permutation of means $\analogy{a}{c}{b}{d}$ deemed intrinsically characteristic of analogy by the Ancients~(cf. \citep{aris_eth}).
The equivalence between the original form \analogy{a}{b}{c}{d\rm{\!}} and these two forms implies the other five.\footnote{
We shall not use this result in the rest of this article,
but we recall that the eight equivalent forms of the analogy establish a correspondence with the group of transformations of the corners of the square,
\ie, the dihedral group, denoted $D_8$.}

As for the analogy by generalized mean,
the inversion of the ratios is given by the symmetry of the equality in $m_p(b, c) = m_p(a, d)$.
The commutativity of addition in $(b^p + c^p)$,
multiplication in $b \times c$ and
in $\min(b, c)$ or $\max(b, c)$
yields the permutation of the means.
The eight equivalent forms of the analogy are therefore met by the analogy of generalized means.

\section{Main results over positive terms}
\label{section:reductions}

We will first consider the case when $a,b,c,d\in \mathbb{R}^*$.
We now show that our definition (Definition~\ref{definition:generalized:analogy}) not only unifies the classical notions of arithmetic and geometric analogy, but also leads to equivalence with an infinite number of analogies in power~$p$.
In other words, there is a canonical form to which we can reduce any analogy in power~$p$ (except for $p=-\infty$ and $p=+\infty$). We can choose, for instance, to reduce to arithmetic analogy.

As we will see in Section~\ref{section:perspectives}, many of these results can in fact be generalised to complex numbers.

\subsection{General case of multiplication by a positive number}
\label{subsection:reduction:multiplication:case:general}

It is easy to see that, for any four positive real numbers,
the analogy in~$p$ between these numbers can be transformed into another analogy of the same power~$p$,
between these same numbers multiplied by any positive number.
That is, we have
\begin{theorem}
$
\label{reduction:multiplication}
\forall p \in \Retoile,
\quad
\forall (a, b, c, d) \in (\Retoileplus)^4, 
$
\begin{equation}
\forall \lambda \in (0,+\infty), \quad
	\analogy[][::^p]{a}{b}{c}{d}
		\Leftrightarrow\;
	\analogy[][::^p]{\lambda a}{\lambda b}{\lambda c}{\lambda d}.
\end{equation}
\end{theorem}

It is easy to see that
this result holds in  case the factor~$\lambda$ is a complex number.
This gives a generalization of analogy to all cases where
the numbers are all on the same line of the complex plane that goes through 0,
and are all on the same side of the line relative to 0.

\begin{remark} Note that, depending on $p$, other invariance properties also hold. For instance, it is easy to see that analogies are invariant under (geometric) translations of the form $x\to x+\lambda$ when $p=1$.    
\end{remark}

\subsection{Special case of division by $d$, \ie, reduction to the unit interval}
\label{subsection:reduction:division:special:case}

From the above property, we deduce that
we can multiply by the inverse of any of the non-zero terms.
This term divided by itself becomes 1.
Consequently, any analogy in the power $p$
can be reduced,
by dividing all terms by one of them,
to an analogy of the same power where one of the terms is 1.

By using the eight equivalent forms of the analogy (see~\ref{subsection:eight:equivalent:forms}), it is always possible to place the largest term in the last position as $d$.
In this case, dividing by $d$ gives an analogy of the same power, where the last term is 1,
and all other terms are less than 1,
\ie,
all terms are in~$]0; 1]$.
As a corollary of Theorem~\ref{reduction:multiplication},
we thus have the following equivalence, in the case where $d$ is greater than or equal to the other terms:
\begin{corollary}
$
\label{reduction:division}
\forall p \in \Retoile,
\;
\forall (a, b, c, d) \in (\Retoileplus)^4, 
\:
a, b, c \leq d
$
\begin{equation}
    \analogy[][::^p]{a}{b}{c}{d}
	   \quad\Leftrightarrow\quad
	\analogy[][::^p]{\frac{a}{d}}{\frac{b}{d}}{\frac{c}{d}}{1}.
\end{equation}
\end{corollary}

\subsection{Reduction to arithmetic analogy}
\label{subsection:reduction:arithmetic}

Any analogy in power $p$, with $p$ different from $-\infty$ or $+\infty$, can be transposed into an arithmetic analogy,
\ie, an analogy in power 1.

In the general case where $p \neq 0$, by taking the power~$p$ of the terms of an analogy,
we obtain the desired equivalence
stated in the following theorem.
For the case $p = 0$, see the Supplementary material.
\begin{theorem}
\label{main:result:reduction:analogy:arithmetic}
$
\forall p \in \Retoile,
\quad
\forall (a, b, c, d) \in (\Retoileplus)^4, 
$
\begin{equation}
	\analogy[][::^p]{a}{b}{c}{d}
		\quad\Leftrightarrow\quad
	\analogy[][::^1]{a^p}{b^p}{c^p}{d^p}.
\end{equation}
\end{theorem}
This constitute the second main result of this article.
On one hand it states that our intuitive understanding about arithmetic analogy, \ie, invariance of ratio $b - a$, applies to any analogy, modulus the transformation by raising to power $p$ or taking the logarithm.
On the other hand, it 

\subsection{Reduction to a canonical form}
\label{subsection:reduction:arithmetic:canonical}

By combining the reduction to the unit interval and the reduction to arithmetic analogy,
we draw the following theorem that states
that,
for a quadruple $(a, b, c, d)$ of positive numbers arranged in ascending order,
if there is an analogy in power $p \neq 0$ between these numbers,
then
this analogy can be reduced to an arithmetic analogy
where one term is necessarily the unit.

\begin{theorem}
$
\label{reduction:canonical:reduction}
\forall p \in \Retoile,
\quad
\forall a < b < c < d \in (\Retoileplus)^4, 
$
\[
	\analogy[][::^p]{a}{b}{c}{d}
		\quad \Leftrightarrow\quad
	\analogy[][::^1]{(\frac{a}{d})^p}{(\frac{b}{d})^p}{(\frac{c}{d})^p}{1}
\]
with the terms ordered as follows if $p$ is positive:
\[
		(\frac{a}{d})^p \;\leq\; (\frac{b}{d})^p \;\leq\; (\frac{c}{d})^p \;\leq\; 1,
\]
and as follows if $p$ is negative:
\[
		(\frac{a}{d})^p \;\geq\; (\frac{b}{d})^p \;\geq\; (\frac{c}{d})^p \;\geq\; 1.
\]
\end{theorem}

\subsection{Reductions across analogies by multiplication of powers}
\label{subsection:reduction:among:analogies}

The reduction to arithmetic analogy, \ie, $p = 1$,
is of course a particular case. 
The generalization of the equivalence with different powers is stated by the following theorem
that shows how one can play with these powers.
One should be however be cautious and note that the powers are non null.
\begin{theorem}
$
\label{reduction:multiplication:of:powers}
\forall (p, q) \in (\Retoile)^2,
\quad
\forall (a, b, c, d) \in (\Retoileplus)^4, 
$
\begin{equation}
	\analogy[][::^{pq}]{a}{b}{c}{d}
		\;\Leftrightarrow\;
	\analogy[][::^p]{a^q}{b^q}{c^q}{d^q}
		\;\Leftrightarrow\;
	\analogy[][::^q]{a^p}{b^p}{c^p}{d^p}.
\end{equation}
\end{theorem}

As a direct consequence, for a given analogy in power $p$,
by rewriting $p = -1 \times -p$,
the analogy in the opposite power is valid for the inverses of the terms.
\[
	\analogy[][::^p]{a}{b}{c}{d}
		\quad \Leftrightarrow\quad
	\analogy[][::^{-p}]{\frac{1}{a}}{\frac{1}{b}}{\frac{1}{c}}{\frac{1}{d}}
\]

\subsection{Existence and uniqueness of $p$ for any quadruple}
\label{subsection:existence:and:uniqueness}

We now examine the existence of an analogy in $p$
for a given quadruple $(a, b, c, d)$ of real numbers.
We assume that these numbers are positive (hence non-zero),
\ie, they are in $(0;+\infty)$.
In addition, importantly,
we assume that they are arranged in ascending order $a < b < c < d$,
with strict inequality.
The following theorem can be established.
Its proof is given in the Supplementary material.

\begin{theorem}
\label{main:result:analogy:reals}
Given four terms $a$, $b$, $c$ and $d$,
positive and ordered in ascending order,
there exist a unique $p$ for which there is an analogy between these terms. In symbols,
$\forall (a, b, c, d) \in \mathbb{R}^4,$
\begin{equation}
	a < b < c < d
		\quad \Rightarrow \quad
	\exists ! p \in \mathbb{R} : \;\;
        \analogy[][::^p]{a}{b}{c}{d}
\label{analogy:reals}
\end{equation}
\end{theorem}

Now, four real numbers, all different, can always be ordered.
Hence, the preceding theorem can be interpreted as follows:
\begin{quote}
{\it Given four positive real numbers,
all different,
we can always see an analogy between them
and this analogy is unique.}
\end{quote}
We can paraphrase:
\begin{quote}{\it 
There is always an angle, and it is unique,
under which we can see an analogy between any four numbers
(positive, all different)
provided we order them in ascending order.}
\end{quote}

The above statements constitute one of the main results in this article.
These statements are somewhat surprising, as one would intuitively assume that there is no reason for the existence of an analogy among any arbitrary four positive real numbers.

\subsection{Calculation in practice}

In practice, the calculation of $p$ for a quadruple of positive numbers can be implemented by a dichotomic search, and $p$ can be calculated to a precision that can be set in advance to serve as a halting criterion.
Of course, this dichotomic search should be launched after checking that the quadruple does not correspond to a particular case of $p$ being zero or infinite.

Figure~\ref{figure:value:p:for:analogy} visualizes how the power $p$ for a given quadruple of positive real numbers is defined.
The solid line gives the values of the generalized means for $a = 2$ and $d = 5$, with $p$ on the x-axis.
The dotted line is the same for $b = 3.5$ and $c = 4.5$.
The value of $p$ for which the analogy $\analogy[][::^p]{a}{b}{c}{d,}$, \ie, $\analogy[][::^p]{2}{3.5}{4.5}{5}$ holds is given by the intersection of the solid line with the dotted line.
In this case $p \simeq 3.06$.

\begin{figure}
\begin{tikzpicture}
\begin{axis}
	[
		xmin = -100, xmax = 100,
		ymin = 1.0, ymax = 6.0,
		xlabel = {$p$},
		ylabel = {generalized mean},
        extra y ticks = {3.5, 4.5},
        extra y tick labels = {$3.5$, $4.5$},
    ]
	\addplot[thick, domain=-100:-0.01] {(0.5 * (2.0 ^ x + 5.0 ^ x)) ^ (1/x)};
	\addplot[thick, domain=0.01:100] {(0.5 * (2.0 ^ x + 5.0 ^ x)) ^ (1/x)} ;
	\addplot[very thick, dotted, domain=-100:-0.01] {(0.5 * (3.5 ^ x + 4.5 ^ x)) ^ (1/x)};
	\addplot[very thick, dotted, domain=0.01:100] {(0.5 * (3.5 ^ x + 4.5 ^ x)) ^ (1/x)} ;

	\addplot[dotted, domain=-100:100] {5} ;
	\addplot[dotted, domain=-100:100] {4.5} ;
	\addplot[dotted, domain=-100:100] {3.5} ;
	\addplot[dotted, domain=-100:100] {2} ;

	\addplot [mark=none, black, dotted] coordinates {(3.06, 0) (3.06, 4.)} ;

    \node[draw] at (3.06, 1.5) {\small $3.06$} ; 

\end{axis}
\end{tikzpicture}
\caption{\label{figure:value:p:for:analogy}%
    Determination of $p$ for the analogy $\analogy[][::^p]{a}{b}{c}{d}$ to hold by taking the intersection of the curves for the generalized means
}
\end{figure}

\section{
Extensions to numerical analogies}
\label{section:remarks}

We now consider the extension to real and complex terms. We start by showing that any analogical equation is solvable over the complex numbers. We also question  the existence of $p$ in this more general setting and provide a visualization of its possible values. We conclude this section with a discussion on the reordering of analogy terms, and comment on particular cases where equality of terms occur or when the terms are Boolean.

\subsection{Solving analogical equations}
\label{section:resolution}

Let $a$, $b$ and $c$ be non-zero positive real numbers arranged in this order and $p$ be any real number,
it is trivial that there is a unique solution to the following equation in the case where $p \neq 0$:
\[
a^p + x^p = b^p + c^p,
\]
or to the following equation corresponding to the case where $p = 0$:
\[
\sqrt{a \times x} = \sqrt{b \times c}.
\]
In other words, there is a unique solution to the analogy equation
\[
\analogy[][::^p]{a}{b}{c}{x}.
\]
Clearly, the same can be said for cases where the unknown $x$ is not in the position of $d$ but in that of $a$ or $b$ or $c$.

For $p = -\infty$ or $+\infty$, different cases should be examined.
In some cases, the solution is unique because it must be equal to the min or the max.
In other cases, any number greater than the min or less than the max will do it, so that the solution is not unique.

It is obvious that the previous statement that there exists a solution to any analogy equation, can be extended to complex numbers, thus removing the constraint on the ordering.
This is our third main result.
\begin{theorem}
\label{main:result:solving:complex}
$\forall p \in R, \forall (a, b, c) \in \mathbb{C} \setminus \{0\},$
\[
\exists x \in \mathbb{C} \setminus \{0\} : \;
\analogy[][::^p]{a}{b}{c}{x}.
\]
\end{theorem}

\subsection{Condition for the existence of $p$}
\label{subsection:not:d:analogy}

According to the eight equivalent forms of analogy,
the two extremes are interchangeable.
The same applies to the two means. 
In addition, the means and the extremes also play the same role by inversion of the ratios.
The proof of the existence of $p$ for the case where the four given numbers (all different) are ordered relies on the possibility that the curves of the generalized means of the extremes and the means intersect.
In other words, $p$ exists if and only if the extremes frame the means, or conversely the means frame the extremes.

\subsection{Visualization of the possible values for $p$}
\label{section:possible:values:p}

\begin{figure}
	\begin{center}
        \includegraphics[height=0.7\columnwidth]{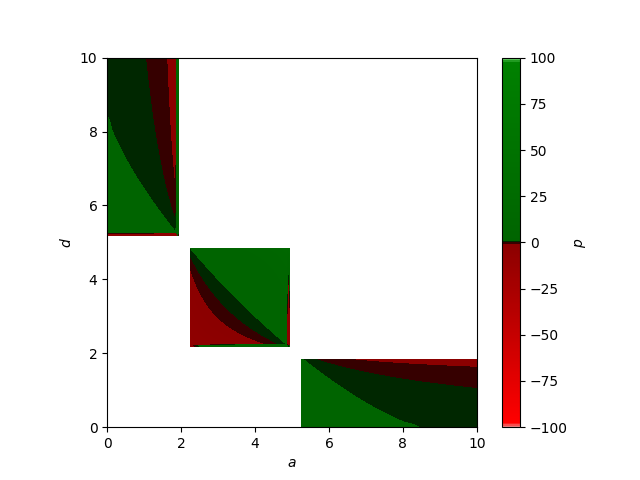}
    \end{center}
	\caption{\label{figure:graphe_axd_de_p}
		Visualization of the values of $p$ for $\analogy[][::^p]{a}{2}{5}{d}$.
	}
\end{figure}

Figure~\ref{figure:graphe_axd_de_p} visualizes the possible values taken by $p$,
when $b$ and $c$ are fixed, here with respective values of $2$ and $5$, and $a$ and $d$ scan the space of possible values,
\ie, three portions of the plane:
the first one for $a < 2$ and $5 < d$,
the second one for $2 < a < 5$ and $2 < d < 5$ and
the third one for $5 < a$ and $d < 2$.
These portions of the plane are explainable thanks to the 8 possible equivalent forms of analogy.
The color scale shows the value of $p$.
The rectangles should touch, but they do not on the picture, due to the limitation of the colors displayed, in the range $[-100, 100]$.
Although not really distinguishable in the picture,
the reader should perceive the hyperbole $a = (2 \times 5) / d$ as
a black curve ($p = 0$).

A similar figure is given in the Supplementary material for the case where $a$ and $b$ are fixed.

\subsection{Reordering any four terms}
\label{subsection:reordering}

In Section~\ref{subsection:existence:and:uniqueness},
we assumed that the four numbers were arranged in ascending order (indeed, we can be satisfied with the extremes framing the means, or vice versa).
Since not every quadruple is ordered, we need to consider the various possibilities.
Elementary combinatorial considerations (24 possibilities), combined with the eight equivalent forms of analogy (24 / 8 = 3),
shows that there are in fact only three reordering possibilities relevant to analogy
(see, \eg~\citep[p.~119]{lep_dhdr_2003}):
\[
\analogy{a}{b}{c}{d}
\quad \text{or} \quad
\analogy{a}{c}{d}{b}
\quad \text{ou} \quad
\analogy{a}{d}{b}{c}.
\]

What was said in Section~\ref{subsection:existence:and:uniqueness}, \ie, the possibility of seeing an analogy between any four numbers, is therefore true for any quadruple not necessarily ordered, provided we specify which reordering is applied.

Note that the case $a = b = c = d$ is very special, as no reordering is required.
In this case, all 24 different entries are possible and equivalent.
We continue our treatment of such cases below.

\subsection{Cases of equality between terms}

With the above remark, we have just encountered a case of equality.
We study them hereafter more thoroughly.

\paragraph{Case of equality $b = c$ only.}
Suppose the terms of the analogy are arranged in the order $a$, $b$, $c$, $d$.
It is possible that $b$ be equal to $c$,
which makes a continuous analogy (see Section~\ref{section:means:generalized}).
In this case, it is clear that, if $a \neq d$ (and $a$ and $d$ are both different from $b$), $p$ is unique and is given by the intersection of the horizontal line $y = b = c$ with the curve of the generalized means of $a$ and $d$.

\paragraph{Case of equality $a = b$ and $c = d$.}
In this case, the two generalized mean curves for $a$ and $d$ and for $b$ and $c$ are superimposed.
The power for analogy is not unique, since any $p$ of $\mathbb{R}$ can be used to write the equality of the mean whatever it may be.
The values $-\infty$ and $+\infty$ are also possible.
Therefore,  $p$ is not unique in this case.

\paragraph{Case of equality $a = d$.}
The terms $a$ and $d$ being the extremes, \ie, the max and min of the terms of the analogy we then necessarily have $a = b = c = d$.
In this case, the
the analogy in~$p$ is true for any~$p$ in~$\mathbb{R}$ and even for $p = -\infty$ or $+\infty$.
There is therefore no uniqueness of $p$ in this case.

In section~\ref{section:reductions}, we divided by $d$, which assumes that it is different from $0$.
If $d$ tends towards 0 and is the maximum of the four positive numbers $a$, $b$, $c$ and $d$,
then in the limit $a = b = c = d = 0$.
As seen above, the analogy will then be true for any value of $p$, and even for $p = 0$,
by defining the value we assign to $0^0$ as a limit.
For the sake of continuity, we would assume in the limit that $0^0 = 0$.

\subsection{Special case of Booleans}

The analogy between Booleans has been presented and studied in several works, such as~\citep{prade_cbrrd_2017,couceiro2,couceiro3}.
Here we establish the link with the analogy in power~$p$.

To establish the link between Booleans and real numbers,
we naturally denote the values false and true by~0 and~1.
There are basically three possible analogies between Booleans:
\[
\analogy{0}{0}{0}{0} 
\quad \text{or} \quad
\analogy{0}{0}{1}{1} 
\quad \text{or} \quad
\analogy{1}{1}{1}{1}
\]
The second analogy can be equivalently rewritten as $\analogy{0}{1}{0}{1}$.

The three analogies above correspond the case $a = b = c = d$ for the first and third ones and the case $a = b$ and $c = d$ for the second one.
It is therefore interesting to note that in all these cases,
as has been seen above for cases of equality, $p$ is not unique,
and all values, including the infinite values, are valid.

But it is also possible to consider the analogy
\[
\analogy{0}{1}{1}{0} 
\]
which is equivalent to $\analogy{1}{0}{0}{1}$ by inversion of ratio (see~\citep{klein_ecai_1982} for a practical occurrence of this case).
These two equivalent forms of the same analogy are explained by considering that the ratio is the logical negation.
The equality of the ratios makes them valid.

However, this analogy is not 
of the model that extends the analogy formulas between sets to Booleans~\citep{lepage_iarml_2023}, nor does it allow a justification based on the minimality of algorithmic complexity~\citep{prade_sqaru_2017}.
Nor does this analogy fit into analogy model in power $p$ proposed in this article, for the reason that the terms are not arranged in ascending order no matter the equivalent forms considered.


\section{Possible extensions and open problems}
\label{section:perspectives}

In this section, we discuss extensions of the previous results by inspecting the possibility of
firstly defining analogy with negative terms and
secondly examining the possibility of having non-real numbers, \ie, complex numbers for powers and terms.

\subsection{Analogy with negative numbers}
\label{subsection:negative:numbers}

Let us start by remarking that $\analogy[][::^1]{-2}{-3}{4}{5}$ cannot be considered a valid analogy because we cannot write: $-2 - (-3) = 4 - 5$, \ie, $1 = -1$.
On the contrary, $\analogy[][::^1]{-3}{-2}{4}{5}$ seems plausible as there is no problem in writing $-3 - (-2) = 4 - 5$,  \ie, $-1 = -1$.
These examples lead us to understand that there exists a relationship between the signs of the terms of an analogy and their ordering.

In the remainder of this section,
the terms $a$, $b$, $c$ and $d$ are positive real numbers.
We propose to extend analogies
with two negative terms based on the above examples.
We state that there should be an exchange of ratio
when taking the inverse
of the two terms in a ratio.
\begin{equation*}
    \analogy[][::^p]{a}{b}{c}{d}
        \;\;\Rightarrow\;\;
    \analogy[][::^p]{(-1\times b)}{(-1\times a)}{c}{d} 
\end{equation*}
Applying the exchange of the means
(see Section~\ref{subsection:eight:equivalent:forms})
implies:
\begin{equation*}
    \analogy[][::^p]{a}{b}{c}{d}
        \;\;\Rightarrow\;\;
    \analogy[][::^p]{(-1\times b)}{c}{(-1 \times a)}{d} 
\end{equation*}
From this,
we can derive a natural extension for four negative numbers.
\begin{equation*}
    \analogy[][::^p]{a}{b}{c}{d}
        \;\;\Rightarrow\;\;
    \analogy[][::^p]{(-1\times a)}{(-1\times b)}{(-1\times c)}{(-1 \times d)} 
\end{equation*}
This natural extension is justified as follows:
\begin{align*}
    \analogy[][::^p]{a}{b}{c}{d}
        & \Rightarrow\;\;
    \analogy[][::^p]{-1\times b}{-1 \times a}{-1\times d}{-1\times c} 
    \\
        & \Rightarrow\;\;
    \analogy[][::^p]{-1\times d}{-1\times c}{-1\times b}{-1 \times a} 
    \\
        & \Rightarrow\;\;
    \analogy[][::^p]{-1\times a}{-1\times b}{-1\times c}{-1 \times d}, 
\end{align*}
where the first implication follows by parallel application of exchange of ratio when taking the inverse of the terms, the second implication follows by symmetry of conformity and the third one by reading inversion (see Section~\ref{subsection:eight:equivalent:forms}).

The only cases with negative numbers that remain to explore are
the case where only one number is negative
and
the case where the means are both negative
while the extremes are both positive
(or the contrary, as the means and the extremes are interchangeable).
These two cases remain open and constitute a topic for future research.

This extension still needs inspection as it requires care when manipulating it.
For instance, 
composition of powers, seen in~\ref{subsection:reduction:among:analogies}, cannot be applied when negative numbers are involved.
Let us consider $\analogy[][::^1]{-3}{-2}{4}{5}$ again.
The following sequence of equivalences leads to a contradiction: a configuration of positive terms where the extremes do not frame the means,
\ie, a configuration that is not an analogy in any real number power.
\begin{align*}
    \analogy[][::^1]{-3}{-2}{4}{5}
    & \Leftrightarrow
        \analogy[][::^{1/2}]{(-3)^2}{(-2)^2}{4^2}{5^2}
    \\
    & \Leftrightarrow
        \analogy[][::^{1/2}]{3^2}{2^2}{4^2}{5^2}
    \\
    & 
    \,\Leftrightarrow \,
        \analogy[][::^1]{3}{2}{4}{5},
\end{align*}
and the latter is clearly false: $\analogy[][\;{::\!\!\!\!\!/}^1\;]{3}{2}{4}{5}$.

\subsection{Analogy with non-real powers or terms}
\label{subsection:non-real:terms:powers}

In Subsection~\ref{subsection:reduction:multiplication:case:general}, we touched upon the possibility of having complex number terms when mentioning the fact that nothing forbids the multiplicative factor $\lambda$ to be complex,
and we mentioned complex numbers for solving analogies in Theorem~\ref{main:result:solving:complex} of Section~\ref{section:resolution}.
Section~\ref{subsection:negative:numbers} led to questions about possibly non-real powers to account for reordering of terms in the case of negative numbers.

Here, we wonder whether it would be possible to capture reordering within analogy.
For four positive real numbers $a$, $b$, $c$ and $d$ in increasing order,
and the unique real number $p$
such that $\analogy[][::^p]{a}{b}{c}{d,}$
what is the type of powers $q$ and $r$ that would allow us to enforce analogy 
for the other two orderings
that we saw in Subsection~\ref{subsection:reordering},
\ie,
$\analogy[][::^q]{a}{c}{d}{b}$
and
$\analogy[][::^r]{a}{d}{b}{c}$\,?
Can they be real numbers or should they be complex numbers or any other kind of numbers?
\begin{align*}
    \left\{
		\begin{aligned}
			\analogy[][::^p]{a}{b}{c}{d} \\
			\analogy[][::^q]{a}{c}{d}{b} \\
			\analogy[][::^r]{a}{d}{b}{c} \\
		\end{aligned}
	\right.
    & \;\;\Leftrightarrow\;\;
    \left\{
		\begin{aligned}
			a^p + d^p &= b^p + c^p \\
			a^q - d^q &= -b^q + c^q \\
			a^r - d^r &= b^r - c^r \\
		\end{aligned}
	\right.
    \\
    & \;\;\Leftrightarrow\;\;
    \left\{
		\begin{aligned}
			a^p + d^p &= b^p + c^p \\
			a^q + (d \times e^{i\pi/q})^q &= (b \times e^{i\pi/q})^q + c^q \\
			a^r + (d \times e^{i\pi/r})^r &= b^r + (b \times e^{i\pi/r})^r. \\
		\end{aligned}
	\right.
\end{align*}
The problem is the existence of such $q$ and $r$
and their possible analytical expression in function of
$p$,
$a$, $b$, $c$ and $d$.
Now, the latter system of equations,
that is explained by recalling that $e^{i\pi} = -1$
(and thus $(e^{i\pi/q})^q = -1$),
would suggest that complex number terms too should be considered to render an account of this enforcement of analogy on the three possible orderings.
We leave this problem open.

\section{Conclusion}
\label{section:conclusion}

The approach taken in this article is a model-theoretic approach to analogy.
We have studied the link between analogy and means, or more precisely, generalized means.
We have leveraged this notion to generalize the notion of analogy on numbers and we defined a notion of analogy in power $p$.
Classical analogies, like arithmetic analogy and geometric analogy, are obviously only special cases for $p=1$ and $p=0$ of analogy in power~$p$.

In particular, we have shown that, given any four positive real numbers, it is always possible to see an analogy between them, thanks to a well-chosen power, which is unique if these numbers are all different.
Put in another way; there is generally a unique perspective, \ie, a power $p$,
under which we can see an analogy between any four positive numbers.
We have also shown that any analogy can be reduced (by bijection) to a canonical form, giving rise to an infinite number of equivalent analogies.
In particular, this shows that all analogies, including classical ones, can be treated by means of an equivalent arithmetic analogy.

This work opens up a number of new avenues.
First of all,
although general results can be obtained not only for four positive numbers,
but also for some mixture of negative and positive real numbers
and even some complex numbers,
the landscape is still incomplete.
Another avenue concerns the semantics of the parameter $p$ which, in a way, summarizes the information implicit in $a, b, c$ and $d$ to their analogy in $p$, \ie, $\analogy[][::^p]{a}{b}{c}{d}$.
This raises the question of whether it is possible to merge the information contained in a quadruple of numbers by simply giving the power of their analogy.

\section{Acknowledgments}
This work is the result of a collaboration made possible by the first author's sabbatical stay with the second author's organization.
The authors would like to thank Waseda University Research Promotion Section and the LORIA International Chair for their funding.

This work benefited from a research grant from the Japan Society for the Promotion of Science, Kiban C, n\textsuperscript{o}~21K12038, entitled
``Theoretically grounded algorithms for the automatic extraction of analogy test sets in natural language processing'' and was partially supported by the ANR project
 ``Analogies: from theory to tools and applications'' (AT2TA), ANR-22-CE23-0023.

\bibliography{ECAI2024_short}

\begin{thebibliography}{53}
\providecommand{\natexlab}[1]{#1}
\providecommand{\url}[1]{\texttt{#1}}
\expandafter\ifx\csname urlstyle\endcsname\relax
  \providecommand{\doi}[1]{doi: #1}\else
  \providecommand{\doi}{doi: \begingroup \urlstyle{rm}\Url}\fi

\bibitem[Allen and Hospedales(2019)]{AllenH19}
C.~Allen and T.~M. Hospedales.
\newblock Analogies explained: Towards understanding word embeddings.
\newblock In \emph{ICML 2019}, volume~97 of \emph{Proceedings of Machine
  Learning Research}, pages 223--231. {PMLR}, 2019.

\bibitem[Alsaidi et~al.(2021)Alsaidi, Decker, Lay, Marquer, Murena, and
  Couceiro]{alsaidiTrans}
S.~Alsaidi, A.~Decker, P.~Lay, E.~Marquer, P.-A. Murena, and M.~Couceiro.
\newblock On the transferability of neural models of morphological analogies.
\newblock In \emph{AIMLAI 2021}, volume~1 of \emph{PKDD/ECML Workshops}, pages
  76--89, Bilbao/Virtual, Spain, 2021.

\bibitem[Aristote(1997)]{aris_eth}
Aristote.
\newblock \emph{{\'E}thique \`a {N}icomaque}.
\newblock Librairie philosophique J.~Vrin, Paris, [1er tirage 1990] edition,
  1997.
\newblock Trad. J.~Tricot.

\bibitem[Badra and Lesot(2022)]{BadraL22}
F.~Badra and M.~Lesot.
\newblock Coat-apc: When analogical proportion-based classification meets
  case-based reasoning.
\newblock In \emph{ATA@ {ICCBR} 2022}, volume 3389 of \emph{{CEUR} Workshop
  Proceedings}, pages 43--56, Nancy, France, 2022.

\bibitem[Badra et~al.(2023)Badra, Lesot, Marquer, and Couceiro]{OursFadi}
F.~Badra, M.-J. Lesot, E.~Marquer, and M.~Couceiro.
\newblock {Some Perspectives on Similarity Learning for Case-Based Reasoning
  and Analogical Transfer}.
\newblock In \emph{IARML@IJCAI 2023}, volume 3492, pages 16--29. CEUR-WS.org,
  2023.

\bibitem[Bitton et~al.(2023)Bitton, Yosef, Strugo, Shahaf, Schwartz, and
  Stanovsky]{DafnaSaaai}
Y.~Bitton, R.~Yosef, E.~Strugo, D.~Shahaf, R.~Schwartz, and G.~Stanovsky.
\newblock {VASR:} visual analogies of situation recognition.
\newblock In \emph{(AAAI 2023) and (IAAI 2023) and (EAAI 2023)}, pages
  241--249, Washington, DC, USA, 2023. {AAAI} Press.

\bibitem[Bouraoui et~al.(2018)Bouraoui, Jameel, and Schockaert]{BouraouiJS18}
Z.~Bouraoui, S.~Jameel, and S.~Schockaert.
\newblock Relation induction in word embeddings revisited.
\newblock In \emph{{COLING} 2018}, pages 1627--1637. Association for
  Computational Linguistics, 2018.

\bibitem[{Chesneau dit Du Marsais} and
  Yvon(1751--1772)]{dumarsais_analogie_1771}
C.~{Chesneau dit Du Marsais} and C.~Yvon.
\newblock Analogie.
\newblock In \emph{Encyclop{\'e}die}. Chez Briasson, Paris, 1751--1772.

\bibitem[Collins and Somers(2003)]{collins_ebmt_2003}
B.~Collins and H.~Somers.
\newblock \emph{Recent Advances in Example-Based Machine Translation}, chapter
  {EBMT} seen as case-based reasoning, pages 115--153.
\newblock Springer Netherlands, Dordrecht, 2003.
\newblock ISBN 978-94-010-0181-6.

\bibitem[Cornu{\'e}jols et~al.(2020)Cornu{\'e}jols, Murena, and
  Olivier]{CornuejolsMO20}
A.~Cornu{\'e}jols, P.-A. Murena, and R.~Olivier.
\newblock Transfer learning by learning projections from target to source.
\newblock In \emph{{IDA} 2020}, volume 12080 of \emph{LNCS}, pages 119--131,
  Springer, 2020.

\bibitem[Couceiro and Lehtonen(2023)]{classification}
M.~Couceiro and E.~Lehtonen.
\newblock {G}alois theory for analogical classifiers.
\newblock \emph{Annals of Mathematics and Artificial Intelligence}, 2023.

\bibitem[Couceiro et~al.(2017)Couceiro, Hug, Prade, and Richard]{couceiro3}
M.~Couceiro, N.~Hug, H.~Prade, and G.~Richard.
\newblock Analogy-preserving functions: A way to extend boolean samples.
\newblock In \emph{IJCAI 2017}, pages 1--7, 2017.

\bibitem[Couceiro et~al.(2018)Couceiro, Hug, Prade, and Richard]{couceiro2}
M.~Couceiro, N.~Hug, H.~Prade, and G.~Richard.
\newblock {Behavior of Analogical Inference w.r.t. Boolean Functions}.
\newblock In \emph{IJCAI 2018}, pages 2057--2063, 2018.

\bibitem[Drozd et~al.(2016)Drozd, Gladkova, and Matsuoka]{DrozdGM16}
A.~Drozd, A.~Gladkova, and S.~Matsuoka.
\newblock Word embeddings, analogies, and machine learning: Beyond king - man +
  woman = queen.
\newblock In \emph{{COLING} 2016}, pages 3519--3530, 2016.

\bibitem[Fahandar and H{\"u}llermeier(2018)]{FahandarH18}
M.~A. Fahandar and E.~H{\"u}llermeier.
\newblock Learning to rank based on analogical reasoning.
\newblock In \emph{{AAAI}-18}, pages 2951--2958, 2018.

\bibitem[Fahandar and H{\"u}llermeier(2021)]{FahandarH21}
M.~A. Fahandar and E.~H{\"u}llermeier.
\newblock Analogical embedding for analogy-based learning to rank.
\newblock In \emph{{IDA} 2021}, volume 12695 of \emph{LNCS}, pages 76--88,
  Springer, 2021.

\bibitem[Fam and Lepage(2024)]{fam_jetai_2022}
R.~Fam and Y.~Lepage.
\newblock Organising lexica into analogical grids: a study of a holistic
  approach for morphological generation under various sizes of data in various
  languages.
\newblock \emph{Journal of Experimental \& Theoretical Artificial
  Intelligence}, 36\penalty0 (1):\penalty0 1--26, 2024.

\bibitem[Gentner(1983)]{gentner_structure_mapping_1983}
D.~Gentner.
\newblock Structure mapping: A theoretical model for analogy.
\newblock \emph{Cognitive Science}, 7\penalty0 (2):\penalty0 155--170, 1983.

\bibitem[Gladkova et~al.(2016)Gladkova, Drozd, and Matsuoka]{GladkovaDM16}
A.~Gladkova, A.~Drozd, and S.~Matsuoka.
\newblock Analogy-based detection of morphological and semantic relations with
  word embeddings: what works and what doesn't.
\newblock In \emph{SRW@HLT-NAACL 2016}, pages 8--15, 2016.

\bibitem[Goel(2019)]{Goel19}
A.~K. Goel.
\newblock Computational design, analogy, and creativity.
\newblock In \emph{Computational Creativity - The Philosophy and Engineering of
  Autonomously Creative Systems}, pages 141--158. Springer, 2019.

\bibitem[H{\"o}lder(1882)]{hoelder_grenzwerthe_1882}
O.~L. H{\"o}lder.
\newblock Grenzwerthe von {R}eihen an der {C}onvergenzgrenze.
\newblock \emph{Math. Annalen}, 20\penalty0 (4):\penalty0 535--549, 1882.

\bibitem[H{\"{u}}llermeier(2020)]{Hullermeier20}
E.~H{\"{u}}llermeier.
\newblock Towards analogy-based explanations in machine learning.
\newblock In \emph{{MDAI} 2020}, volume 12256 of \emph{LNCS}, pages 205--217,
  Springer, 2020.

\bibitem[Jarnac et~al.(2023)Jarnac, Couceiro, and Monnin]{JarnacCM23}
L.~Jarnac, M.~Couceiro, and P.~Monnin.
\newblock Relevant entity selection: Knowledge graph bootstrapping via
  zero-shot analogical pruning.
\newblock In \emph{{CIKM} 2023}, pages 934--944. {ACM}, 2023.

\bibitem[Keane and Smyth(2020)]{KeaneS20}
M.~T. Keane and B.~Smyth.
\newblock Good counterfactuals and where to find them: {A} case-based technique
  for generating counterfactuals for explainable {AI} {(XAI)}.
\newblock In \emph{{ICCBR} 2020}, volume 12311 of \emph{LNCS}, pages 163--178,
  Springer, 2020.

\bibitem[Klein(1982)]{klein_ecai_1982}
S.~Klein.
\newblock Culture, mysticism and social structure and the calculation of
  behavior.
\newblock In \emph{{ECAI 1982}}, pages 141--146, 1982.

\bibitem[Langlais et~al.(2009)Langlais, Zweigenbaum, and Yvon]{lan_eacl_09}
P.~Langlais, P.~Zweigenbaum, and F.~Yvon.
\newblock Improvements in analogical learning: application to translating
  multi-terms of the medical domain.
\newblock In \emph{{EACL}~2009}, pages 487--495, Athens, Greece, 2009.

\bibitem[Lepage(2003)]{lep_dhdr_2003}
Y.~Lepage.
\newblock \emph{De l'analogie rendant compte de l'analogie en linguistique}.
\newblock M{\'e}moire d'abilitation {\`a} diriger les recherches, Grenoble
  University, May 2003.

\bibitem[Lepage(2023)]{lepage_iarml_2023}
Y.~Lepage.
\newblock Formulae for the solution of an analogical equation between
  {B}ooleans using the {S}heffer stroke ({NAND}) or the {P}ierce arrow ({NOR}).
\newblock In \emph{{IARML@IJCAI 2023}}, pages 3--14, August 2023.

\bibitem[Lieber et~al.(2021)Lieber, Nauer, and Prade]{LieberNP21}
J.~Lieber, E.~Nauer, and H.~Prade.
\newblock When revision-based case adaptation meets analogical extrapolation.
\newblock In \emph{{ICCBR} 2021}, volume 12877 of \emph{LNCS}, pages 156--170,
  Springer, 2021.

\bibitem[Linzen(2016)]{Linzen16}
T.~Linzen.
\newblock Issues in evaluating semantic spaces using word analogies.
\newblock In \emph{RepEval@ACL 2016}, pages 13--18, 2016.

\bibitem[Michel(1949)]{michel_persee_1949}
P.-H. Michel.
\newblock Les m{\'e}di{\'e}t{\'e}s.
\newblock \emph{Revue d'histoire des sciences et de leurs applications},
  2\penalty0 (2):\penalty0 139--178, 1949.

\bibitem[Mikolov et~al.(2013)Mikolov, Yih, and Zweig]{MikolovYZ13}
T.~Mikolov, W.~Yih, and G.~Zweig.
\newblock Linguistic regularities in continuous space word representations.
\newblock In \emph{{HLT-NAACL} 2013}, pages 746--751, 2013.

\bibitem[Mitchell(2021)]{Mitchell21}
M.~Mitchell.
\newblock Abstraction and analogy-making in artificial intelligence.
\newblock \emph{Annals of the New York Academy of Sciences}, 1505\penalty0
  (1):\penalty0 79--101, 2021.

\bibitem[Murena et~al.(2020)Murena, Al-Ghossein, Dessalles, and
  Cornu{\'e}jols]{murena_ijcai_2020}
P.-A. Murena, M.~Al-Ghossein, J.-L. Dessalles, and A.~Cornu{\'e}jols.
\newblock Solving analogies on words based on minimal complexity
  transformation.
\newblock In \emph{{IJCAI-20}}, pages 1848--1854, 7 2020.
\newblock Main track.

\bibitem[Nagao(1984)]{nag_ex_eng}
M.~Nagao.
\newblock A framework of a mechanical translation between {J}apanese and
  {E}nglish by analogy principle.
\newblock In \emph{Int. NATO symposimu on Artificial and human intelligence},
  pages 173--180. Elsevier Science Publishers, NATO, 1984.

\bibitem[Petersen and van~der Plas(2023)]{PetersenP23}
M.~R. Petersen and L.~van~der Plas.
\newblock Can language models learn analogical reasoning? {I}nvestigating
  training objectives and comparisons to human performance.
\newblock In \emph{EMNLP 2023}, pages 16414--16425, 2023.

\bibitem[Peyre et~al.(2019)Peyre, Laptev, Schmid, and Sivic]{PeyreSLS19}
J.~Peyre, I.~Laptev, C.~Schmid, and J.~Sivic.
\newblock Detecting unseen visual relations using analogies.
\newblock In \emph{{IEEE ICCV} 2019}, pages 1981--1990, 2019.

\bibitem[Prade and Richard(2017{\natexlab{a}})]{prade_cbrrd_2017}
H.~Prade and G.~Richard.
\newblock Analogical proportions and analogical reasoning - an introduction.
\newblock In \emph{Case-Based Reasoning Research and Development}, pages
  16--32, Cham, 2017{\natexlab{a}}. Springer.

\bibitem[Prade and Richard(2017{\natexlab{b}})]{prade_sqaru_2017}
H.~Prade and G.~Richard.
\newblock Boolean analogical proportions - axiomatics and algorithmic
  complexity issues.
\newblock In \emph{ECSQARU}, pages 10--21, Cham, 2017{\natexlab{b}}.

\bibitem[{Rallier des Ourmes}(1751--1772)]{rallier_proportion_1771}
J.~J. {Rallier des Ourmes}.
\newblock Proportion.
\newblock In \emph{Encyclop{\'e}die}. Chez Briasson, Paris, 1751--1772.

\bibitem[Rhouma and Langlais(2014)]{rhouma-langlais:2014:Coling}
R.~Rhouma and P.~Langlais.
\newblock Fourteen light tasks for comparing analogical and phrase-based
  machine translation.
\newblock In \emph{{COLING} 2014}, volume Technical Papers, pages 444--454,
  Dublin, Ireland, August 2014.

\bibitem[Rhouma and Langlais(2018)]{rhouma_iccbr_2018}
R.~Rhouma and P.~Langlais.
\newblock Experiments in learning to solve formal analogical equations.
\newblock In \emph{{ICCBR}-18)}, pages 438--453. Springer, 2018.

\bibitem[Sadeghi et~al.(2015)Sadeghi, Zitnick, and Farhadi]{SadeghiZF15}
F.~Sadeghi, C.~L. Zitnick, and A.~Farhadi.
\newblock Visalogy: Answering visual analogy questions.
\newblock In \emph{NIPS 2015}, pages 1882--1890, 2015.

\bibitem[Schluter(2018)]{Schluter18}
N.~Schluter.
\newblock The word analogy testing caveat.
\newblock In \emph{NAACL-HLT 2018, Volume 2 (Short Papers)}, pages 242--246,
  2018.

\bibitem[Schnabel et~al.(2015)Schnabel, Labutov, Mimno, and
  Joachims]{SchnabelLMJ15}
T.~Schnabel, I.~Labutov, D.~M. Mimno, and T.~Joachims.
\newblock Evaluation methods for unsupervised word embeddings.
\newblock In \emph{{EMNLP} 2015}, pages 298--307, 2015.

\bibitem[Trouillon et~al.(2016)Trouillon, Welbl, Riedel, Gaussier, and
  Bouchard]{TrouillonWRGB16}
T.~Trouillon, J.~Welbl, S.~Riedel, {\'{E}}.~Gaussier, and G.~Bouchard.
\newblock Complex embeddings for simple link prediction.
\newblock In \emph{{ICML} 2016}, volume~48 of \emph{{JMLR}}, pages 2071--2080,
  2016.

\bibitem[Turney(2006)]{turney_compling_06}
P.~D. Turney.
\newblock Similarity of semantic relations.
\newblock \emph{Computational Linguistics}, 32\penalty0 (2):\penalty0 379--416,
  2006.

\bibitem[Ushio et~al.(2021)Ushio, Anke, Schockaert, and
  Camacho{-}Collados]{UshioASC20}
A.~Ushio, L.~E. Anke, S.~Schockaert, and J.~Camacho{-}Collados.
\newblock {BERT} is to {NLP} what alexnet is to {CV:} can pre-trained language
  models identify analogies?
\newblock In \emph{ACL/IJCNLP 2021}, volume~1, pages 3609--3624, 2021.

\bibitem[Wang et~al.(2021)Wang, Wang, Yang, Zhang, Chen, and Qi]{WangWYZCQ21}
M.~Wang, S.~Wang, H.~Yang, Z.~Zhang, X.~Chen, and G.~Qi.
\newblock Is visual context really helpful for knowledge graph? {A}
  representation learning perspective.
\newblock In \emph{{MM} 2021}, pages 2735--2743. {ACM}, 2021.

\bibitem[Wang et~al.(2019)Wang, Li, Li, and Zeng]{WangLLZ19}
Z.~Wang, L.~Li, Q.~Li, and D.~Zeng.
\newblock Multimodal data enhanced representation learning for knowledge
  graphs.
\newblock In \emph{{IJCNN} 2019}, pages 1--8. {IEEE}, 2019.

\bibitem[Xie et~al.(2017)Xie, Liu, Luan, and Sun]{XieLLS17}
R.~Xie, Z.~Liu, H.~Luan, and M.~Sun.
\newblock Image-embodied knowledge representation learning.
\newblock In \emph{{IJCAI} 2017}, pages 3140--3146. ijcai.org, 2017.

\bibitem[Zervakis et~al.(2022)Zervakis, Vincent, Couceiro, Schoenauer, and
  Marquer]{TSV}
G.~Zervakis, E.~Vincent, M.~Couceiro, M.~Schoenauer, and E.~Marquer.
\newblock An analogy based approach for solving target sense verification.
\newblock In \emph{NLPIR 2022}, pages 144--151. {ACM}, 2022.

\bibitem[Zhang et~al.(2023)Zhang, Li, Chen, Liang, Deng, and Chen]{LDC23}
N.~Zhang, L.~Li, X.~Chen, X.~Liang, S.~Deng, and H.~Chen.
\newblock Multimodal analogical reasoning over knowledge graphs.
\newblock In \emph{{ICLR} 2023}, 2023.

\end{thebibliography}

\appendix

\section{Supplementary material}

\subsection{Proofs for reflexivity, symmetry and transitivity of conformity in $p$}

For the reflexivity of $::^p$,
it is trivial to note that,
for any two positive numbers $a$ and $b$,
for any $p$,
we always have
\begin{equation}
\label{eq:reflexivity:conformity:supplementary}
    \analogy[][::^p]{\graybox{$a$}}{\graybox{$b$}}{a}{b},
\end{equation}
i.e.,
\begin{equation}
\label{eq:reflexivity:conformity:explanation:supplementary}
	\moygen{r}{p}{\graybox{$a$}}{b}
		=
	\moygen{r}{p}{\graybox{$b$}}{a}.
\end{equation}

We visualize by shading the fact that,  
while
the first $a$
de~\eqref{eq:reflexivity:conformity:supplementary}
corresponds
to the first $a$
of~\eqref{eq:reflexivity:conformity:explanation:supplementary},
on the other hand
the first $b$
to the left of the conformity in $p$
in~\eqref{eq:reflexivity:conformity:supplementary}
corresponds
the second $b$
to the right of the equal sign
in~\eqref{eq:reflexivity:conformity:explanation}.

The symmetry of $::^p$ is expressed as follows:
\[
	\analogy[][::^p]{a}{b}{c}{d}
		\Leftrightarrow
	\analogy[][::^p]{c}{d}{a}{b}.
\]
It is verified because:
\begin{align*}
	\analogy[][::^p]{a}{b}{c}{d}
	& \Leftrightarrow
		m_p(a, d)
			=
		m_p(b, c)
    \\
	& \Leftrightarrow
		m_p(b, c)
			=
		m_p(a, d)
    \\
	& \Leftrightarrow
		m_p(c, b)
			=
		m_p(d, a)
    \\
	& \Leftrightarrow
		\analogy[][::^p]{c}{d}{a}{b}.
\end{align*}

Let us repeat that transitivity is not generally necessary for analogy.
It just happens to exist for $::^p$.
This can be stated as:
\begin{equation}
	\left\{
		\begin{aligned}
			\analogy{a}{b}{c}{d} \\
			\analogy{c}{d}{e}{f}
		\end{aligned}
	\right.
		\quad\Rightarrow\quad
	\analogy{a}{b}{e}{f}.
\end{equation}

For the general case where $p$ is in $(-\infty; +\infty)$,
and different from 0,
we have:
\begin{align*}
	\left\{
		\begin{aligned}
			m_p(a, d) = m_p(b, c) \\
			m_p(c, f) = m_p(d, e) 
		\end{aligned}
	\right.
	&
		\quad\Leftrightarrow\quad
	\left\{
		\begin{aligned}
			a^p - b^p = c^p - d^p \\
			c^p - d^p = e^p - f^p 
		\end{aligned}
	\right.
	\\
	&
		\quad\Rightarrow\quad
			a^p - b^p = e^p - f^p 
	\\[1ex]
	&
		\quad\Rightarrow\quad
			m_p(a, f) = m_p(b, e).
\end{align*}

In the case where $p$ tends to zero, we know that $m_0(a, d) = \sqrt{a \times d}.$
It is therefore easy to obtain, provided that none of the terms is zero:
\begin{align*}
	\left\{
		\begin{aligned}
			m_0(a, d) = m_0(b, c) \\
			m_0(b, c) = m_0(e, f) 
		\end{aligned}
	\right.
	&
		\quad\Leftrightarrow\quad
	\left\{
		\begin{aligned}
			\sqrt{a / b} = \sqrt{c / d} \\ 
			\sqrt{c / d} = \sqrt{e / f} 
		\end{aligned}
	\right.
	\\[1ex]
	&
		\quad\Rightarrow\quad
			\sqrt{a / b} = \sqrt{e / f} 
 	\\
	&
		\quad\Leftrightarrow\quad
			m_0(a, f) = m_0(b, e).
\end{align*}

In the case where $p$ tends towards minus infinity,
we would like to do the following reasoning, which is blocked at the point mentioned below by the absence of implication.
\begin{align*}
	\left\{
		\begin{aligned}
			m_{-\infty}(a, d) = m_{-\infty}(b, c) \\
			m_{-\infty}(c, f) = m_{-\infty}(d, e) 
		\end{aligned}
	\right.
	&
		\quad\Leftrightarrow\quad
	\left\{
		\begin{aligned}
			\min(a, d) = \min(b, c) \\
			\min(c, f) = \min(d, e) 
		\end{aligned}
	\right.
	\\[1ex]
	&
		\quad\not\Rightarrow\quad
			\min(a, f) = \min(b, e) 
 	\\[1ex]
	&
		\quad\Leftrightarrow\quad
			m_{-\infty}(a, f) = m_{-\infty}(b, e).
\end{align*}

In general, we do not have the implication.
But if we assume that $a$, $b$, $c$, $d$ are arranged in ascending order and that $e$ and $f$ are greater than or equal to $d$,
then we can show transitivity.
The same applies to plus infinity, for which we simply need to replace $\min$ by $\max$ and reverse the direction of the inequalities.
and reverse the direction of the inequalities.

\subsection{Proofs for reductions}

\subsubsection{General case of multiplication by a positive number}

It is indeed easy to show that, first in the case where~$p$ is different from~0:

$\analogy[][::^p]{a}{b}{c}{d}$
\begin{align*}
    \qquad
		& \quad\Leftrightarrow\quad
			\moyp{p}{a}{d} = \moyp{p}{b}{c} \\
		& \quad\Leftrightarrow\quad
			\lambda \times \moyp{p}{a}{d} = \lambda \times \moyp{p}{b}{c} \\
		& \quad\Leftrightarrow\quad
			\moyp{p}{(\lambda a)}{(\lambda d)} = \moyp{p}{(\lambda b)}{(\lambda c)} \\
		& \quad\Leftrightarrow\quad
			\analogy[][::^p]{\lambda a}{\lambda b}{\lambda c}{\lambda d}.
\end{align*}

In the case where $p$ is equal to 0,
it is also easy to show that:
\begin{align*}
	\analogy[][::^0]{a}{b}{c}{d}
		& \quad\Leftrightarrow\quad
			\sqrt{ad} = \sqrt{bc} \\
		& \quad\Leftrightarrow\quad
			\lambda \times \sqrt{ad} = \lambda \times \sqrt{bc} \\
		& \quad\Leftrightarrow\quad
			\sqrt{\lambda^2ad} = \sqrt{\lambda^2bc} \\
		& \quad\Leftrightarrow\quad
			\sqrt{(\lambda a)(\lambda d)} = \sqrt{(\lambda b)(\lambda c)} \\
		& \quad\Leftrightarrow\quad
			\analogy[][::^0]{\lambda a}{\lambda b}{\lambda c}{\lambda d}.
\end{align*}

In the case where $p$ tends to $-\infty$,
we have:
\begin{align*}
	\analogy[][::^{-\infty}]{a}{b}{c}{d}
		& \quad\Leftrightarrow\quad
			\min(a, d) = \min(b, c) \\
		& \quad\Leftrightarrow\quad
			\min(\lambda a, \lambda d) = \min(\lambda b, \lambda c) \\\
		& \quad\Leftrightarrow\quad
			\analogy[][::^{-\infty}]{\lambda a}{\lambda b}{\lambda c}{\lambda d}.
\end{align*}

And similarly for $p$ tending to $+\infty$,
replacing $min$ by $max$.

\subsubsection{Reduction to arithmetic analogy}

This is easily proved as follows:
\begin{align*}
	\analogy[][::^p]{a}{b}{c}{d}
	&
		\quad\Leftrightarrow\quad
			\frac{1}{2}(a^p + d^p) = \frac{1}{2}(b^p + c^p)
	\\
	&
		\quad\Leftrightarrow\quad
			\analogy[][::^1]{a^p}{b^p}{c^p}{d^p}.
\end{align*}
Another way of doing this
is to simply write the following equivalences,
bearing in mind that the analogy to the power of 1 is
the arithmetic analogy where
ratio is subtraction and
conformity is equality:
\begin{align*}
	\analogy[][::^p]{a}{b}{c}{d}
	&
		\quad\Leftrightarrow\quad
			a^p - b^p = c^p - d^p
	\\
	&
		\quad\Leftrightarrow\quad
			\analogy[][::^1]{a^p}{b^p}{c^p}{d^p}.
\end{align*}

In the case where $p=0$,
the transformation to be applied to pass from the geometric analogy to the arithmetic analogy
is to take the logarithm, \ie:
\begin{align*}
	\analogy[][::^0]{a}{b}{c}{d}
	&
		\quad\Leftrightarrow\quad
			\sqrt{a \times d} = \sqrt{b \times c}
	\\
	&
		\quad\Leftrightarrow\quad
			\frac{1}{2}(\ln a + \ln d) = \frac{1}{2}(\ln b + \ln c)
    \\
	&
		\quad\Leftrightarrow\quad
			\analogy[][::^1]{\ln a}{\ln b}{\ln c}{\ln p}.
\end{align*}

The property does not apply to $-\infty$ and $+\infty$.
There is no transformation allowing us to move to an arithmetic analogy from the simple equality of the $\min$ or $\max$ of extremes and means.

\subsection{Existence and uniqueness of $p$}

We want to prove Theorem~\ref{main:result:analogy:reals}.
\begin{align*}
 \forall (a, b, c, d) \in \mathbb{R}^4, &
 \\
	a < b < c < d &
		\;\; \Rightarrow \;\;
	\exists ! p \in \mathbb{R} :
        \analogy[][::^p]{a}{b}{c}{d}
\end{align*}
in other words,
\begin{equation}
\label{equality::generalized:means}
	\exists ! p \in \mathbb{R} \;\; \colon \;\; 
	\moygen{r}{p}{a}{d}
		=
	\moygen{r}{p}{b}{c}.
\end{equation}

\subsubsection{Existence of $p$}
\label{subsection:existence}

Consider the following function in~$p$:\[
\delta(p)
=
\
\moygen{r}{p}{a}{d}
-
\moygen{r}{p}{b}{c}.
\]
This function is simply the difference between the left and right sides of the equation~\eqref{equality::generalized:means}.

The proof of the existence of some $p$ is carried out in two stages.
The first stage of the demonstration
consists in showing that the function
is continuous and strictly increasing.
The second step
is to show that
the extreme values of this function are one negative, the other one positive.
The proof just consists in saying that,
the function being
continuous
and
bounded between a negative and a positive number,
it has at least one value at which it cancels out.


First, we establish that $\delta$ is continuous.
The function $\delta$ being the difference of two functions $m_p(a, d)$ and $m_p(b, c)$
both continuous on $\mathbb{R}$, is continuous on $\mathbb{R}$.
At 0, the limit is $\sqrt{ad} - \sqrt{bc}$.


Let us now look at the extreme values of $\delta$.
Let us consider the case where $p$ tends towards $-\infty$.
Recall that $a$, $b$, $c$ and $d$ are assumed to be ordered in a strictly increasing fashion.
The function $\delta$ therefore tends to the value $\min(a, d) - \min(b, c) = a - b$.
Now, since $a$ is less than $b$, the limit value in $-\infty$ is negative:
\[
\lim_{p \rightarrow -\infty} \delta(p) = a - b < 0.
\]

For the case where $p$ tends towards $+\infty$,
the same reasoning applies, replacing $\min$ by $\max$:
$\max(a, d) - \max(b, c) = d - c$.
The value obtained, $d - c$, is positive because $c < d$:
\[
\lim_{p \rightarrow +\infty} \delta(p) = d - c > 0.
\]


The fact that $\delta$ is continuous and lies between two values
one negative on the left,
the other one positive on the right,
implies by Cauchy's intermediate value theorem that $\delta$ cancels out,
in other words, there is at least one value of~$p$ defining an analogy between the four terms $a$, $b$, $c$ and~$d$.

\subsubsection{Uniqueness of $p$}
\label{subsection:uniqueness:p}

First, we show that if there are two analogies, then there are in fact three.
Let $p$ and $q$ be the abscissas of two points of intersection of the curves of the generalized means of $a, d$ and $b, c$. Since $a$ and $d$ enclose $b$ and $c$,
the curve for $b, c$ passes at $p$ below that of $a, d$ coming from $-\infty$ and above it at $q$ coming from $+\infty$.
It is therefore below and above between the two values $p$ and $q$.
This means that there exists some $r$ between $p$ and $q$ for a third point of intersection.
Either $r$ is negative and we have two negative values $p$ and $r$; or $r$ is positive and we have two positive values $r$ and $q$.
We show below that these two cases lead to contradiction.

Suppose there are two values \textcolor{black}{$0 < r < q$} such that the analogy holds.
We can restrict ourselves to the case $0 < a < b < c < 1$
according to the result seen in Section~\ref{subsection:reduction:division:special:case}.
Then we have:
\begin{align*}
	\analogy[][::^r]{a}{b}{c}{1}
		& \Leftrightarrow
			\textcolor{black}{b^r + c^r - 1 = a^r}
	\\[2ex]
	\analogy[][::^{q}]{a}{b}{c}{1}
		& \Leftrightarrow
			b^{q} + c^{q} - 1 = a^{q}
	\\
		& \Leftrightarrow
			 b^{q} + c^{q} - 1 = \textcolor{black}{a^r} \times a^{q-r}
	\\
		& \Leftrightarrow
			 b^{q} + c^{q} - 1 = \textcolor{black}{(b^r + c^r - 1)} \times a^{q-r}  
	\\
		& \Leftrightarrow
			 \underbrace{\frac{1 - ( b^{q} + c^{q} ) }{1 - ( b^{r} + c^{r} ) }}_\text{$> 1$} =  \underbrace{a^{q-r}}_\text{$< 1$}
\end{align*}

The last line is justified as follows.
On the one hand $b^{q} + c^{q} < b^{p} + c^{p}$ because $b < c < 1$ and $0 < r < q$, hence, a ratio greater than 1.
On the other hand $a^{q-r} < 1$ because $a< 1$ and $q - r >0$.
This is a contradiction:
hence, there cannot be two distinct values $q$ and $r$.

For two negative values $p < r < 0$, the same demonstration applies using the equivalent analogy seen in Section~\ref{subsection:reduction:among:analogies}: $\analogy[][::^{-p}]{1/a}{1/b}{1/c}{1/d.}$

For the sake of completeness, as a special case, asking whether $p = 0$ is simply a matter of checking the uniqueness of the case $bc = ad$. 

Figure~\ref{figure:value:p:for:analogy} visualizes the uniqueness of $p$ for particular values of $a$, $b$, $c$ and $d$.

\subsection{Visualization of the values of $p$}
\label{subsection:visualization::values:p}

\begin{figure}
	\begin{center}
		\includegraphics[height=0.6\columnwidth]{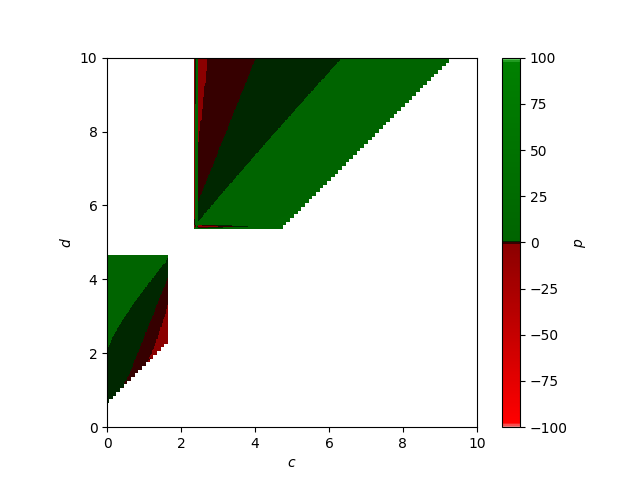}
	\end{center}
	\caption{\label{figure:graphe_cxd_de_p}
		Visualization of the values of $p$ for $\analogy[][::^p]{2}{5}{c}{d}$.
	}
\end{figure}

Figure~\ref{figure:graphe_cxd_de_p} is the same kind of visualization as Figure~\ref{figure:graphe_axd_de_p},
but when $a$ and $b$ are fixed, with values of $2$ and $5$.
The axes are thus for $c$ and $d$.
We draw the values of $p$ for $c < d$.
Observe that $p$  exists in only two portions of the half-plane:
the first one for $c < 2$ and $d < 5$,
the second one for $c > 2$ and $d > 5$.
The same remarks about the imperfections of the picture apply here
as for Figure~\ref{figure:graphe_axd_de_p}.

\end{document}